# Empowering machine learning models with contextual knowledge for enhancing the detection of eating disorders in social media posts

José Alberto Benítez-Andrades [a,*], María Teresa García-Ordás [b], Mayra Russo [c], Ahmad Sakor [c], Luis Daniel Fernandes Rotger [d] and Maria-Esther Vidal [c]

[a] *SALBIS Research Group, Department of Electric, Systems and Automatics Engineering, Universidad de León, León, Spain*
*E-mail: jbena@unileon.es*

[b] *SECOMUCI Research Group, Escuela de Ingenierías Industrial e Informática, Universidad de León,León, Spain*
*E-mail: mgaro@unileon.es*

[c] *Leibniz University of Hannover and L3S Research Center and TIB Leibniz Information Centre for Science and Technology, Germany*
*E-mails: mrusso@l3s.de, ahmad.sakor@tib.eu, maria.vidal@tib.eu*

[d] *Bakken and Baeck GmbH, Germany*
*E-mail: luis@bakkenbaeck.no*

**Abstract.** Social networks have become information dissemination channels, where announcements are posted frequently; they also serve as frameworks for debates in various areas (e.g., scientific, political, and social). In particular, in the health area, social networks represent a channel to communicate and disseminate novel treatments' success; they also allow ordinary people to express their concerns about a disease or disorder. The Artificial Intelligence (AI) community has developed analytical methods to uncover and predict patterns from posts that enable it to explain news about a particular topic, e.g., mental disorders expressed as eating disorders or depression. Albeit potentially rich while expressing an idea or concern, posts are presented as short texts, preventing, thus, AI models from accurately encoding these posts' contextual knowledge. We propose a hybrid approach where knowledge encoded in community-maintained knowledge graphs (e.g., Wikidata) is combined with deep learning to categorize social media posts using existing classification models. The proposed approach resorts to state-of-the-art named entity recognizers and linkers (e.g., Falcon 2.0) to extract entities in short posts and link them to concepts in knowledge graphs. Then, knowledge graph embeddings (KGEs) are utilized to compute latent representations of the extracted entities, which result in vector representations of the posts that encode these entities' contextual knowledge extracted from the knowledge graphs. These KGEs are combined with contextualized word embeddings (e.g., BERT) to generate a context-based representation of the posts that empower prediction models. We apply our proposed approach in the health domain to detect whether a publication is related to an eating disorder (e.g., anorexia or bulimia) and uncover concepts within the discourse that could help healthcare providers diagnose this type of mental disorder. We evaluate our approach on a dataset of 2,000 tweets about eating disorders.

*Corresponding author. E-mail: jbena@unileon.es.

Our experimental results suggest that combining contextual knowledge encoded in word embeddings with the one built from knowledge graphs increases the reliability of the predictive models. The ambition is that the proposed method can support health domain experts in discovering patterns that may forecast a mental disorder, enhancing early detection and more precise diagnosis towards personalized medicine.

## 1. Introduction

The COVID-19 pandemic has considerably burdened mental diseases all over the world [60], and eating disorders (EDs) are not an exception [16,46,62,81]. EDs are health conditions severely disturbing eating behaviors and related thoughts and emotions. A recent study by Zipfel et al. [81] reveals that EDs increased during the pandemic by 15.3% in 2020 with respect to previous years. Moreover, outcomes from a systematic literature review by McLean et al. [43] uncover that children and adolescents are the most vulnerable groups impacted by EDs during the COVID-19 pandemic. This burden of mental incidences raises awareness of the need for early detection mechanisms to effectively take action in clinical and healthcare services. More importantly, these studies provide evidence of the urgent need for scalable methods for effectively supporting communities increasingly suffering from mental disorders.

Social media is increasingly used as a dissemination channel to announce novel treatments or conditions to respond to health-related problems and even to discuss natural disasters [64]. Moreover, social media networks are utilized to promote or prevent the administration of certain interventions, e.g., in mental conditions like eating disorders [9,39]. Furthermore, Artificial Intelligence (AI) has gained momentum in the healthcare field [8,80], and AI-based solutions have been developed for disease prevention [17], pathology detection [73], and treatment prescription [57]. Analyzing the discourse on social networks such as Twitter can help to find answers to relevant problems by applying various ML techniques. Specifically, in mental health, predictive models have successfully been applied [12,63]. Exemplary contributions include the detection of depression [29], suicidal mental tendencies [21], bipolar disorders [24] and other mental disorders [20]. Moreover, the models proposed by [38] can uncover patterns of anorexia in datasets obtained from social networks. These results put in perspective the role of automatic classification in the scalable and effective detection of patterns of communities suffering from mental disorders [75].

Knowledge graphs (KGs), and Semantic Web technologies in general, have been accepted as data structures that enable the natural representation and management of the convergence of data and knowledge [22]. The information contained in knowledge graphs is increasingly used in the scientific community to solve different problems [1,3,67]. Specifically, community-maintained knowledge graphs such as Wikidata [71] or DBpedia [33] represent rich sources of structured knowledge not only from general domains but also in biomedicine [10,30,79]. Figure 1(b) depicts the contextual knowledge obtained through structured data in a portion of Wikidata. We define this contextual knowledge as entities related to a given resource within the knowledge graph, e.g., the Wikidata resource Q254327 represents the concept "anorexia", that is "symptom" and "physiological condition" with "decreased appetite". This contextual knowledge can be interpretable by both humans and machines. This symbolic depiction of anorexia can be encoded in a subsymbolic representation using embeddings computed by Knowledge Graph Embedding methods [19,50,68]. Obtaining knowledge graph embeddings (KGEs) allows for calculating distance or similarity metrics and discovering relatedness among the entities that compose a knowledge graph (shown at the bottom of Fig. 1(b)). This information is of interest to obtain what is known as *semantic enrichment* in different textual problems for which only textual information is initially available. Capturing contextual knowledge via semantic enrichment provides the basis for precise and accurate predictive models. This is of paramount relevance for ensuring the impact and credibility of the ED patterns detected from social media posts using AI methods.

Predictive models can also be built over unstructured data, Bidirectional Encoder Representations from Transformers (BERT) models [15] are exemplary solutions to this problem [2]. Contrary to other models (e.g., BiLSTM-based-Bidirectional Long Short-Memory), BERT models can learn from words in all positions, i.e., from the whole sentence. Figure 1(a) illustrates an example of contextual knowledge extracted from unstructured information for the term "anorexia". Here, this contextual knowledge is based on the information around the word anorexia within a given sentence. These models usually receive texts and labels as input to achieve predictive models that classify texts. By applying this model, vector embeddings are also obtained, but in this case, the encoded contextual content encodes unstructured data (lower part of Fig. 1(a)). BERT models have exhibited high performance in almost any prediction problem where contextual knowledge is extracted from text. Nevertheless, albeit the large number of covered domains and high-quality predictions, BERT models may perform poorly over short texts [35,78].

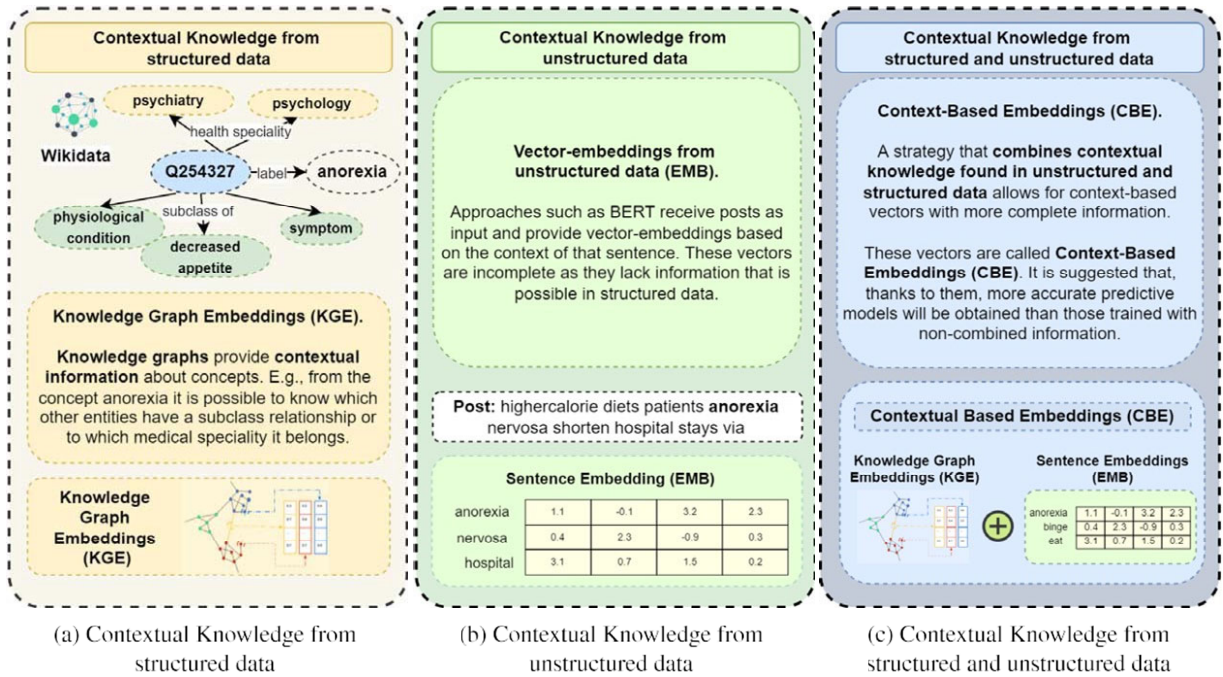

Fig. 1. Contextual knowledge (CK). (a) CK is represented in knowledge graphs and encoded in vector embeddings KGE methods. (b) CK is extracted from unstructured data, e.g., social media posts; it corresponds to the words around a particular concept. (c) The proposed approach, CK extracted from both structured and unstructured data and encoded in context-based embeddings (CBE).

**Problem Statement and Objectives.** This paper addresses the problem of effectively classifying short posts referring to EDs. Our main research objective is to generate vector-embedding representations that encode contextual knowledge reported in structured data structures from knowledge graphs and unstructured corpora (e.g., scientific publications or social media posts). Concretely, we aim at encoding richer contextual knowledge, as depicted in Fig. 1(c). As a result, our goal is to generate vector-embeddings obtained from structured and unstructured data. The result is called contextual-based embeddings (CBEs). We hypothesize that CBEs will empower machine learning models in a way that the accuracy of pattern detection from social media posts is enhanced.

**Proposed Solution.** We propose a hybrid framework that combines vector embeddings generated from methods that extract contextual knowledge from various types of data sources (Fig. 1(c)). As a proof of concept, we show the benefits of the proposed approach while combining word embeddings from pre-trained BERT models and knowledge graph embedding methods learned from Wikidata. Although our approach is domain-agnostic, we will illustrate its performance in the problem of classifying Tweets according to mentions of eating disorders.

The resulting embeddings are used in various state-of-the-art predictive models. The methodology followed to implement the proposed approach is composed of the following steps: **1)** *Contextual-based knowledge from unstructured data*: Vector-embeddings are obtained from unstructured data by applying pre-trained BERT models. **2)** *Contextual-based knowledge from structured data*: RDF2Vec [56] is utilized to traverse the KGs, e.g., Wikidata. The resources to be traversed are extracted from social posts using Named Entity Recognition (NER) and Named Entity Linking (NEL). Thus, an entity embedding encodes the contextual knowledge represented in the traversed neighborhood of the entity which is recognized in the post. **3)** *Combined contextual knowledge*: Contextual-based embeddings (CBEs) are computed by combining BERT and RDF2Vec embeddings. **4)** *Creation of Predictive Models*: CBEs are utilized to train and test the predictive models in a given classification task. The corpus of social media posts, annotated with labels, is also provided as input. In this paper, we have evaluated the proposed framework in a corpus of 2,000 Tweets related to eating disorders. This corpus has been created, curated, and annotated by the authors. The quality of the combined embeddings has been empirically evaluated in 12 state-of-the-art predictive models. The observed results indicate that considering the CBE increases the performance of all the models in accuracy and F-measure. Albeit limited to one corpus, these experimental outcomes provide evidence of the impact that contextual knowledge has on the accuracy of classification tasks conducted by existing methods.

**Contributions.** In summary, the scientific contributions of this work are as follows:

- **A corpus of Tweets related to Eating Disorders**. The dataset was collected using Twitter's API and tagged in four categories. The corpus is composed of 2,000 tweets from which a total of 1,358 different entities are recognized. For these entities, their sentence embedding vectors from text data are calculated using Word2Vec, Tokenisers, and BERT and knowledge graph embeddings are generated using RDF2Vec [56]. The dataset quality is analyzed in terms of diversity and potential bias.
- **A Contextual-based Method to Empower Predictive Models**. The proposed method computes vector embeddings from structured and unstructured data and generates embeddings that integrate the two perspectives. This method is applied to the corpus of Tweets to solve four classification tasks related to eating disorders.
- **Experimental Assessment of the Proposed Methods**. An empirical evaluation is performed on the corpus over 12 different predictive models. Four binary classification tasks are solved. The results indicate that models that use contextual knowledge from structured and unstructured data in combination perform better in 97.5% of the cases, an improvement of up to 15%. The dataset and the code with the testbeds are available in a public repository, thus, ensuring reproducibility [7].

The remainder of this paper is organized as follows. Section 2 presents the fundamentals of RDF, knowledge graphs, knowledge graph embeddings, entity recognition, and entity linking. Section 3 analyzes related approaches from the state of the art. Then Section 4 defines our problem statement and states our proposed solution. This section also describes the main components of the architecture that implements our proposed method. Section 5 reports on the experimental evaluation and discusses the observed outcomes. Finally, we close the paper in Section 6 with conclusions and an overview of future work.

## 2. Preliminaries

### 2.1. Wikidata – Community-maintained knowledge graph

In this research, we use community-maintained knowledge graphs and, more specifically, Wikidata, to obtain information related to the concepts contained in short texts. The Wikidata knowledge graph is stored internally in JSON format and can be edited by any user thanks to the web interface of this knowledge graph. Wikidata can be downloaded in RDF, however, a subject-property-object triple is annotated with qualifiers [41], representing metadata about the triple. Lastly, although Wikidata is a source of encyclopedic knowledge, Waagmeester et al. [72] report the main characteristics of Wikidata as a relevant provider of knowledge from Life Sciences.

### 2.2. Entity recognition and entity linking, making use of knowledge graphs

In our approach, we use named entity recognition and linking tools to extract the entities found in the short texts used as input data and link them to the concepts contained in Wikidata. A named entity linker implements the tasks of named entity recognition (NER) and named entity linking (NEL), allowing the identification of entities in a text and their corresponding resources in a knowledge graph or controlled vocabulary. Although our approach is named entity linker agnostic, as a proof of concept, we make use of the EntityLinker[1] contained in the spaCy

---
[1] https://spacy.io/api/entitylinker

[26] Python library and Falcon 2.0 [58]. Both tools resort to background knowledge to perform the linking process. EntityLinker's knowledge base is updated with Wikidata resources from 2019,[2] while Falcon 2.0 has a background knowledge updated to 2021. The two systems are used because the recognized entities were not always the same, and it was observed that a union of the entities obtained using both tools is more complete.

*2.3. Knowledge graph embeddings*

Knowledge graph embeddings (KGE) are low-dimensional representations of the entities and relationships that make up a knowledge graph. KGEs make possible a generalizable representation of an entity based on its context on a global knowledge graph, in our case, Wikidata. This context allows inferring relationships between concepts. The KGEs provide us with information about the relationships between different terms related to our problem. For example, in an ED context, KGEs provide us with the relationships between 'green tea', 'day', 'good', 'anorexic', 'binge-eating' and 'nah'. Thanks to the magnitude of information contained in Wikidata, the KGEs provide information about the interactions between the concepts contained in the short texts we want to classify.

This information, added to the information found in the tweets themselves, is of vital importance to improve the predictive models that classify these short texts. There are many configurations and methods for calculating these knowledge graph embeddings. In our approach, we use of RDF2Vec [56]. Inspired by Word2vec [44] which represents words in vector space, RDF2Vec applies this method within a knowledge graph. RDF2Vec allows receiving different ways to create sequences of RDF nodes that are then used as input for the Word2vec algorithm. One of the most commonly used strategies is random walks in an RDF graph.

*2.4. Extracting knowledge graph embeddings through RDF2Vec*

RDF2Vec, as proposed by Ristoski et al. [56], adapts the language modeling approach in Word2vec for latent representations of entities in RDF graphs. Ristoski et al. [56] demonstrate that projecting such latent representations of entities into a lower dimensional feature space shows that semantically similar entities appear closer to each other in the knowledge graph where these entities are represented. The following steps are proposed to generate RDF2Vec embeddings from a corpus of short texts:

1. The entities of each of the texts in the dataset are obtained, as explained in Section 2.2.
2. The set of unique entities is used to generate a vocabulary.
3. An extension of RDF2Vec (implemented by the authors) is used to generate knowledge graph embeddings of a set of entities. This procedure calculates the embeddings based on the entities that are in the neighborhood of each entity within Wikidata. This enables us to obtain word vectors based on the contextual knowledge within the entities found in the collected dataset.

*2.5. Strategies for combining embeddings*

In the proposed approach, once the knowledge graph embeddings are obtained, it is necessary to combine them, as the number of entities in each tweet may not always be the same and, therefore, the number of KGEs may also be different. As can be seen in the scientific literature, there are many methods of combining embeddings [4,34,52,54], the simplest of which is to find the average of the total number of embeddings obtained in each tweet. However, this does not seem to be the best strategy. In our approach, we have used smooth inverse frequency (SIF) [4]. SIF also considers the most meaningful words within a sentence; it scores from 0 to 5 pairs of sentences, in our case tweets, according to the similarity between them. SIF is used as it is one of the most widely used strategies [49,76].

---

[2]https://www.kaggle.com/kenshoresearch/kensho-derived-wikimedia-data

## 3. Related work

### 3.1. Deep learning in social media data

The use of social media data to train predictive models that use machine learning and deep learning techniques is becoming increasingly common in the scientific field. A search for the terms "social media data AND deep learning" in Google Scholar returns approximately 225,000 results since 2017.[3] Some of these studies focus on generating predictive models that are able to classify texts within a given domain, such as mental disorder [38], emotions [51], or fake news [77]. These studies resort to machine learning and deep learning techniques, comparing the results obtained and the computational cost of each. Currently, the techniques that obtain the best results in terms of performance and accuracy of the classification models are long short-term memory (LSTM), bidirectional long short-term memory (Bi-LSTM) neural networks, and, since they appeared, bidirectional encoder representations from transformers (BERT) models. Although the predictive models obtained in many studies have a fairly high performance in the form of hit rates, our hypothesis is that most of them could be improved if, in addition, the semantic information of these texts were used.

### 3.2. Knowledge acquisition from knowledge graphs

Knowledge graphs such as Wikidata [71] and DBpedia [33] are increasingly used in research. Moreover, studies have shown that the information they contain, is useful and reliable for use in some fields like the Life Sciences [72]. One of the most important problems scientists face in acquiring knowledge through these collaborative knowledge networks is what is known as Entity Linking. In order to obtain a resource that represents the concept "help" from Wikidata, it is necessary to know whether we are referring to the concept of help as cooperation between people[4] or whether we are referring to the studio album called "Help!" by The Beatles.[5] Due to the growing use of these knowledge graphs for research, tools have emerged that make it possible to obtain, in a simple way, not only the entities in a sentence, but also the link to these concepts in collaborative knowledge graphs. Some examples are the Falcon 2.0 tool [58] and the EntityLinker[6] function from the spaCy [26] library that is implemented and available for use in Python. Thanks to these tools, obtaining information associated with the concepts contained in texts is easier and faster. The extracted linked entities can be used to perform different operations, for example, to calculate the corresponding embeddings for these entities. These calculations can be done using methods such as RDF2Vec [56].

### 3.3. Combining knowledge graphs and deep learning

The contextual knowledge found in knowledge graphs combined with the use of deep learning techniques is being used to solve different research problems. For example, thanks to the semantic information obtained through these knowledge graphs, it is possible to improve the interpretability and explainability of different predictive models powered by deep learning techniques [36]. Improving the interpretability and explainability of predictive models is important, especially when deployed in domains related to health or education. Moreover, knowledge graphs are also being used to help predict relationships between different concepts, such as, the gut microbiota and mental disorders [18] Notwithstanding, there is a lack of studies that have used the contextual knowledge extracted from knowledge graphs combined with the contextual knowledge extracted from unstructured data to classify texts by making use of this information. Our study aims to demonstrate that incorporating semantic enrichment to texts can contribute to improve the performance of predictive models.

---

[3]https://scholar.google.es/scholar?as_ylo=2017&q=social+media+data+AND+deep+learning&hl=es&as_sdt=0,5
[4]https://www.wikidata.org/wiki/Q1643184
[5]https://www.wikidata.org/wiki/Q201816
[6]https://spacy.io/api/entitylinker

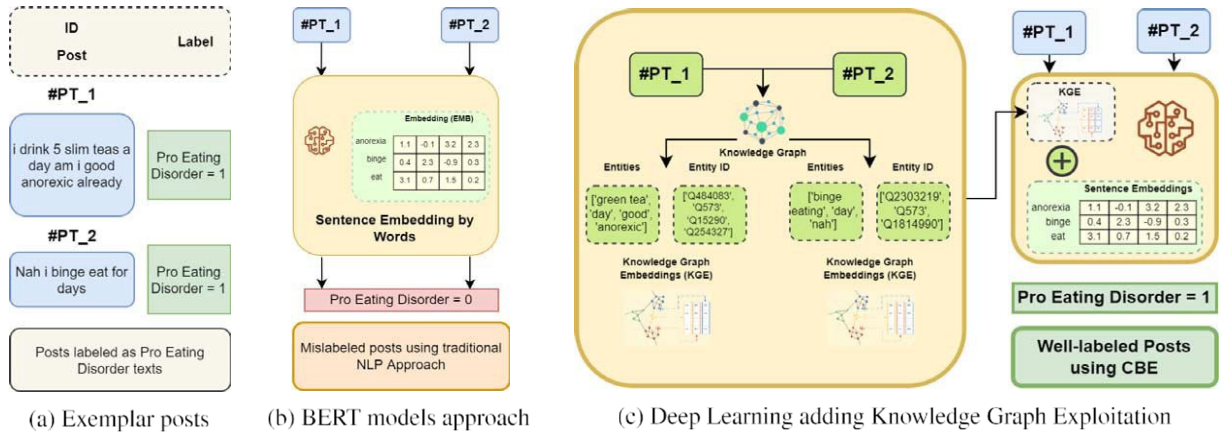

Fig. 2. Motivating example. (a) 2 labeled posts, (b) a text classification model using BERT models approach, (c) short text classification using the novel approach making use of context-based embeddings (CBE).

## 4. Combining contextual knowledge from unstructured and structured data

In this research, we address the problem of effectively classifying short text. We propose a hybrid method based on the combination of contextual knowledge from structured data and unstructured data, resulting in contextual based embeddings. We elaborate below on the problem statement and the proposed solution.

**Problem Statement**. Machine learning models have shown to underperform when applied to short text classification problems. As depicted in Fig. 2, NLP-based approaches, for example, those that incorporate BERT models, continue to demonstrate accuracy problems, and are capable of mislabeling short text posts in binary classification tasks. This problem needs to be addressed as datasets elaborated from short posts obtained from social media, e.g., Twitter, are widely used.

**Proposed Solution**. The approach proposed in this work resorts to contextual knowledge extracted from unstructured data combined with contextual knowledge extracted from structured data to improve the performance of predictive models for short text classification. Vector-embeddings are obtained based on the context of a sentence (unstructured data); secondly, knowledge graph embeddings are obtained (e.g., using RDF2Vec), which represent the contextual knowledge of structured data. Figure 3 depicts the proposed architecture of the empowered model; concisely, it receives as input a corpus composed of short texts and the corresponding entities extracted from knowledge graphs. Both of these are transformed into embedding vectors to be combined. The output is the classification generated by a predictive model after a process of context-based embeddings encoding and decoding.

**Component 1: Context-Based Embeddings Encoding.**
This component has three main objectives: (i) to obtain contextual knowledge through unstructured data (sentence embeddings), (ii) to obtain contextual knowledge through structured data (knowledge graph embeddings), and (iii) to combine this textual knowledge into a single vector structure through the combination of both (context-based embeddings). Objectives (i) and (ii) are achieved through modules that are executed in parallel. Objective (iii) is attained once the two previous modules have been executed. The properties of the three modules that make up this component are detailed below.

**Unstructured based data.** This module generates a vector-embedding for each word within a post. This vector represents the similarity of the terms within the corpus of posts. For example, the term "anorexia" within the sentence *"higher-calorie diets patients anorexia nervosa shorten hospital stays via"* is contextually related to the terms around it. A word-embedding encodes the similarity values of the terms within the corpus of posts. Once all the word-embeddings of a sentence are computed, a vector-embedding is generated; it represents the complete sentence, combining all the vectors of a sentence. This vector-embedding contains contextual knowledge of the posts. Thus, the closeness between terms is encoded, e.g., "anorexia" and "diets". These vectors can be obtained using, e.g.,

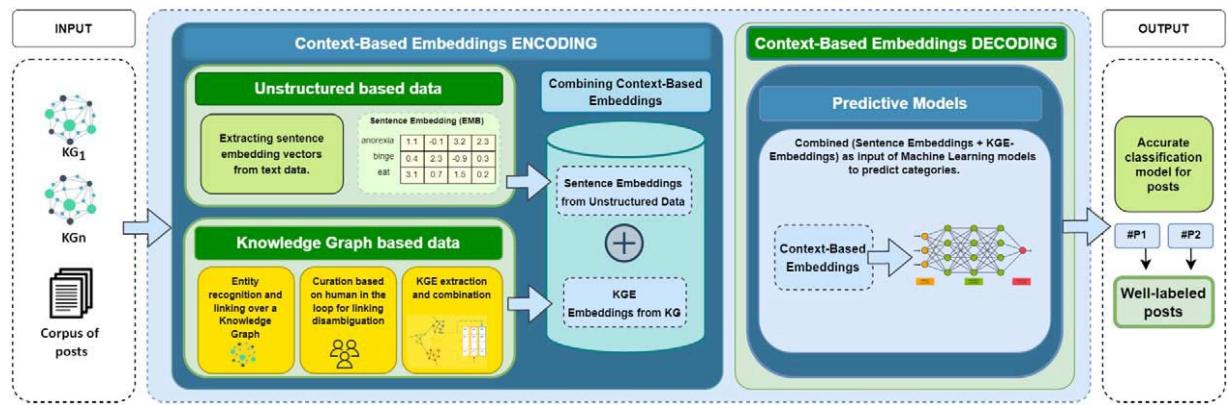

Fig. 3. Architecture of the proposed approach. It receives a corpus of posts and knowledge graphs. The first component, called context-based embeddings encoding, represents how unstructured based data is combined with the information obtained through knowledge graphs applying entity recognition and linking to concepts within a knowledge graph. The knowledge graph-based data module shows the process of knowledge graph embeddings (KGEs) extraction from a knowledge base. Then, sentence embeddings from unstructured data are combined with KGE. The second component, called context-based embeddings, is computed, and predictive machine learning models are executed with these embeddings in decoding, resulting in accurate classification models for posts.

Tokenisers, Word2Vec or BERT. Although this combined vector represents the words of a sentence, it does not include knowledge modeled in structured data.

**Knowledge Graph-based data.** The main objective of this module is to obtain knowledge graph embeddings for each post; these vectors will enrich the ones generated from the unstructured data. Three steps are performed:

*(I) Entity recognition and linking over a Knowledge Graph*. The first step consists of recognizing the entities contained in the short texts used as input and linking the recognized entities to a knowledge graph; these engines are known as named entity recognizers and linkers (e.g., Falcon 2.0 [58] and EntityLinker in the spaCy [26] Python literature). In the current version of this module, the recognized entities are linked to Wikidata. The Falcon 2.0 API[7] extracts the entities of a sentence, as well as the resources that correspond to these entities in Wikidata and DBpedia. The following example illustrates the tasks of entity recognition and linking over the following sentence:

*higher-calorie diets patients anorexia nervosa shorten hospital stays via*

The recognized entities are:

['kilocalorie','diet','patient','anorexia','nervosa']

And the identifiers of these entities in Wikidata are:

['Q26708069', 'Q474191', 'Q181600', 'Q254327', 'Q131749']

*(II) Curation based on human in the loop for linking disambiguation*. Entity recognizers and linkers may be inaccurate. For example, the word "Help" can represent the action of help or support to someone, or it could be a music album called "Help!". In order to assist this type of disambiguation, we resort to a list of *tabu types* to help determine when the linked resource needs to be manually validated and curated. This list includes the following types: *Album, Book, Streets, Organization, Song, and Movie*. This process enhances the quality of the description of a post in terms of entities in a knowledge graph.

*(III) KGE – Embeddings Extraction and Combination*. Knowledge graphs contain contextual knowledge of entities about concepts. In this step, knowledge graph embeddings are obtained for each entity of each post present. For example, the concept "anorexia" is represented in Wikidata as resource Q254327, which is a "symptom" and "physical condition" with "decreased appetite". The symbolic representation of anorexia in Wikidata is encoded in a sub-symbolic representation or knowledge graph embedding. RDF2Vec [56] or SDM-RDF2Vec [61] can generate these vectors. After obtaining each knowledge graph embedding for a post's entities, a combination of these vectors, which have the same size, is generated. As a result, a knowledge graph embedding for each post is generated.

---

[7]https://labs.tib.eu/falcon/falcon2/api-use

**Combining Context-Based Embeddings.** Once the contextual knowledge has been obtained from the unstructured based data and knowledge graphs data modules, this third module is in charge of combining both vectors. Combining these two vectors allows the complete contextual knowledge of each concept to be represented by a single vector. For example, in the initial sentence, we used "highercalorie diets patients anorexia nervosa shorten hospital stays via" we would have a single vector called context-based embedding. This vector would contain the information of the post obtained from its context as unstructured data, but it would also contain knowledge of the neighborhood of the words that compose it. The vector would represent the similarity between each post in the corpus from two different contextual knowledge. This combined vector contains the complete contextual information of each post.

**Component 2: Context-Based Embeddings Decoding.**

This component uses the information contained in the context-based embeddings generated by the first component by decoding them. According to the hypothesis put forward in this study, the complete contextual knowledge contained in these vectors should provide predictive models with a higher degree of accuracy. This knowledge allows the vectors to contain a more accurate inter-post similarity function than vectors generated from unstructured data alone or from unstructured data. In the running example, "anorexia" was close to "nervosa" in the vector obtained from the unstructured data, but not close to "symptom". The combination of both types of embeddings encodes different contextual knowledge; it offers a contextual-based enhanced vector.

**Predictive models.** In this module, different machine learning models are trained. The result is empowered models that outperform traditional models, i.e., have higher accuracy, when classifying a corpus of text. This decoding methodology can be applied not only to text classification problems, but also to other supervised and unsupervised learning problems.

## 5. Experiments and results

The experiments conducted in this study demonstrate that machine learning models enhanced with contextual knowledge can improve accuracy in predictive short text classification models. The objective of our particular case study is to improve the classification of eating disorders in social media posts. The empirical study sets out to answer the following research questions:

**RQ1)** What is the impact of contextual knowledge extracted from structured data on the performance of the model?

**RQ2)** Do BERT pre-trained models based on short texts perform better than other pre-trained models?

**RQ3)** What type of contextual information provides the most accurate predictive models – structured, unstructured or a combination of both?

Data availability of the experiments and code of the empirical evaluation of the proposed approach[8] are publicly available; this facilitates the reproducibility of the reported results.

### 5.1. Experimental setup

The following settings are configured to answer our research questions.

**Benchmarks.** The experiments are executed using a proprietary eating disorders dataset. The dataset was elaborated by collecting tweets from Twitter and contains 2,000 posts labeled as detailed in the Section 5.3. The dataset is divided into training and test sets to train the different predictive models. This separation is performed using a stratified separation function, in a proportion of 70% for training and 30% for testing. This way, it is possible to ensure that the categories are balanced in both sets and to avoid possible overfitting or under fitting of the models.

---

[8] https://github.com/knowledgeb/Combining-Knowledge-Graphs-and-Deep-Learning-techniques-for-Categorizing-Tweets

Table 1
Formulas of the evaluation metrics used

| Metric | Formula |
| --- | --- |
| Precision | $P(c) = \frac{TP}{TP+FP}$ |
| Recall | $R(c) = \frac{TP}{TP+FN}$ |
| $F_1$-score | $F_1 = \frac{2*P*R}{P+R}$ |
| Accuracy | $Acc = \frac{TP+TN}{(TP+TN+FP+FN)}$ |

*Cross-validation.* A k-fold cross-validation has been carried out with 10 folds for all experiments to avoid randomness. Every model has been evaluated through an extensive hyperparameter grid search. The results show the score for the hyperparameter configuration with the best mean value of the ten folds.

**Metrics.** To evaluate the results obtained by each of the models, two different metrics have been used: $F_1$-score ($F_1$) and accuracy (acc) (Table 1). $F_1$-score is a metric used to calculate the effectiveness of a classifier by taking into account its accuracy and recall values. F1 assumes that the two metrics used are equally important in calculating effectiveness. If one of them is more important, a different formula $F_\beta$ would have to be used. The formula used to calculate this metric is as follows, where P equals the precision value and R equals the recall value. Accuracy refers to how close the result of a measurement is to the true value. In statistical terms, accuracy is related to the bias of an estimate. It is represented as the proportion of true results (both true positives (TP) and true negatives (TN)) divided by the total number of cases examined (true positives, false positives, true negatives, false negatives).

**Implementations.** The proposed approach is implemented in Python 3.6. having CUDA 11.1 installed. The following libraries are required in the indicated versions: numpy v1.19.5, params v0.9.0, pandas v1.2.2, pyrdf2vec v0.2.3, qwikidata v0.4.0, spacy v3.4.1, tensorflow v2.4.0, torch v1.10.2, tweet-preprocessor v0.6.0, scikit_learn v0.24.2 and simpletransformers v0.60.9. The experiments are executed on a Windows 10 Pro x64 machine with an Intel® Core® i7-9700K CPU @ 3.60 GHz (eight physical cores, eight threads) and 32 GiB DDR4 RAM with a graphic card NVIDIA GeForce RTX 2080 SUPER (8 GB GDDR6).

**Engines.** Falcon 2.0 [58] and EntityLinker [26] are used to perform entity recognition and linking. Moreover, knowledge graph embeddings are computed with pyRDF2vec [69]; it is configured as follows: *RandomWalker* with *max_depth=4* and *max_walks=50*.

**Models.** 12 different machine learning models are executed to classify the four binary categories. The classification models used are: Random Forest Classifier [13], LSTM [27], Bi-LSTM [28], CNN, CNN+ LSTM [80] and 7 pre-trained BERT models [15]. The pre-trained BERT models are as follows: TweetBERT [45], BERT [14], RoBERTa [37], DistilBERT [59], CamemBERT [42], Albert [31], and FlauBERT [32]. These models receive three inputs: (i) the texts of the posts, (ii) the structured information obtained from knowledge graphs, and (iii) the combination of structured and unstructured information in the form of context-based embeddings (CBE). These models are evaluated using the metrics explained above.

### 5.2. Experimental methodology

The different steps taken to validate the model proposed in this research using a dataset of our own are detailed below. The working pipeline is divided into 4 phases:

1. **Collection and labeling of ED data:** in this phase, texts are collected through web scraping on the social network Twitter. After this collection, the tweets are labeled in the four binary categories of study.
2. **Analysis of the data collected:** once the data has been tagged, an analysis of the texts obtained is carried out. This analysis provides information on the context of the texts obtained, and also on the importance of emojis in the dataset. This analysis allows us to choose pre-processing strategies to avoid possible biases in the experiments.
3. **Knowledge acquisition through knowledge graphs:** once the labeled dataset is curated, information is obtained through knowledge graphs, in this case, Wikidata. In this step, the knowledge graph embeddings of the texts are obtained using RDF2Vec [56]. Then the work of combining these embeddings with the embeddings obtained in the texts is carried out, giving rise to context-based embeddings.

4. **Application of machine learning models:** in this phase, different machine learning models are applied using the texts, the knowledge graph embeddings, and the context-based embeddings as input. In this way, it is possible to compare the accuracy results of these models using traditional approaches and the approach we propose in this research. In addition, to verify that our model is not dependent on the dataset used, an analysis has been carried out on the input dataset and on the output obtained after applying the different models.

The phases mentioned above are explained in detail in the following subsections.

### 5.3. Collecting and labeling ED dataset

Figure 4 depicts the process followed to collect and label the tweets that compose our corpus.

We employ information-filtering-based methods of social media posts using a set of keywords as search queries. This method of collection allows for faster and less costly access to a bigger volume of data and with a specialized level of detail as to the desired output when compared to what could be obtained through traditional collection techniques such as clinical screening surveys [48,74]. Nonetheless, the datasets produced this way come with their limitations due in great part to the pervasive noise associated with their source, which may impact the quality of the data itself, as well as the reliability and representativeness of the results obtained [5,48].

Our chosen social media platform is Twitter, and we perform the capture of tweets using the T-Hoarder tool [11]. The list of hashtags used to filter tweets is anorexia, anorexic, dietary disorders, inappetence, feeding disorder, food problem, binge-eating, eating disorders, bulimia, food issues, and loss of appetite. These hashtags were selected taking into account the most relevant keywords that yielded tweets about eating disorders in other research [23,25, 40,65]. However, this sub-selection could still bias the sampling and results, as we might be making assumptions about the behavior of the users generating the content or overlooking relevant information outside these limits [48]. The collection yielded the capture of 494,025 tweets, and to mitigate issues associated with lexical and semantic redundancy of the content [48], a process of cleaning (removal of duplicates, re-tweets, re-shared content), manual curation and annotation were performed, thus, resulting in the creation of a subset of 2,000 tweets.

To annotate the tweets, we first determine four binary classification problems of interest guided by preexisting ED research [25,40,65]. The resulting categories for each problem are detailed below, followed by the labeling criteria of each tweet:

- **ED I:** Tweets written by people with an ED (1) or not (0). For each tweet, the author's profile was accessed. All those who indicated in their biography to suffer from an ED and/or wrote tweets explicitly indicating that they suffered from an ED were put into the positive class (1).
- **ED II:** Tweets that promote having ED (1) or not (0). In the community, there are terms that expose a person to be in favor of having an ED, for example: proana (proanorexia) or promia (probulimia) [70]. Tweets labeled as promoting EDs were written by authors expressing these terms in their biographies or timelines.
- **ED III:** Tweets of an informative nature (1) or not (0). Tweets that contained information on an ED-related subject were put into the positive class (1). Content that reflected opinions was put in the negative class (0).

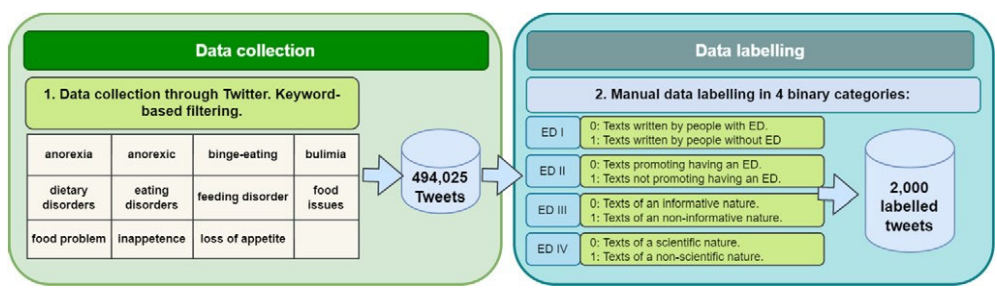

Fig. 4. Data collection and data labeling steps. First, a set of tweets were collected through a keyword-based search. Subsequently, a subset of the tweets was manually labeled into four binary categories.

- **ED IV:** Tweets of scientific nature (1) or not (0). Tweets written by ED health professionals and researchers disseminating relevant information, data, and results were categorized as scientific. Tweets linked to an article in a scientific journal were also categorized as scientific.

This dataset is available for download from github [7].

### 5.4. Dataset analysis

As per the characteristics of the data and the research itself, it is not our objective to profile the sampled users, thus we avoid making assumptions based on 'gender', 'nationality', 'socio-economic background', 'age', to name a few features, that could lead us to fall back on perpetuating damaging stereotypes associated with persons who suffer eating disorders, as recent research has shown that they are global illnesses that do not discriminate [47,53]. However, all the collected tweets belong to users that express themselves in English, a result of the predetermined search criteria. Additionally, as part of the 2,000 tweets, 1,567 belong to unique Twitter accounts, meaning that 433 tweets are concentrated in 183 Twitter accounts, a figure that could indicate there is no presence of over-representation of 'noisy' social media users, notwithstanding the possibility that this distribution includes accounts that belong to nonhumans (i.e., bots, corporate accounts), multiple users posting from the same account, or the same entity posting from different accounts [6,48].

Through an exploration of the frequency of terms associated with the hashtags collected, we managed to identify a predominant mention of anorexia and other derivative terms such as "anorexia", "anorexic", "proana", "ana", "anatwt", "anorexiatips", (1,034 associated hashtags), while an underrepresentation of hashtags associated with other eating disorders, such as bulimia (182 associated hashtags) is identified. This pattern is also replicated in the analysis of the content of the tweets themselves, as the most frequent terms were also associated with anorexia (more than 428 unigram tokens) over bulimia (109 tokens), a small caveat regarding this will be that binge (282 tokens) eating, is a behavior that could both be identified as a symptom of bulimia or a disorder in itself. While this detangling has not been the focus of this study, the results of this exploration leave the door open to perform further captures that could cover a wider variety of eating disorders or for a fine-grained analysis that could help semantically differentiate these disorders in short text analysis. Figure 5 presents a summary of the data analysis. Most frequent terms with unigram tokens (see Fig. 5(b) and Figure 5(a)) and a breakdown of the annotated categories and the corresponding class count is showed in Fig. 5(c). The top 35 hashtags show a predominant mention of 'anorexia' and other derivative terms such as "anorexia", "anorexic", "proana", "ana", "anatwt", "anorexiatips".

Emojis were eliminated from the tweets in the corpus. The following analysis provides evidence of the lack of effect of the emojis' removal.

- The frequencies of the different emojis in the texts were obtained, and it was determined that only 17.9% of texts contained some emoji, i.e., 359 of the 2,000 total texts.
- Statistical calculations were carried out on the distributions of emojis according to the four categorizations made. To statistically analyze these data, two statistical techniques were used: (i) overlap analysis and (ii) Spearman's rho correlation.

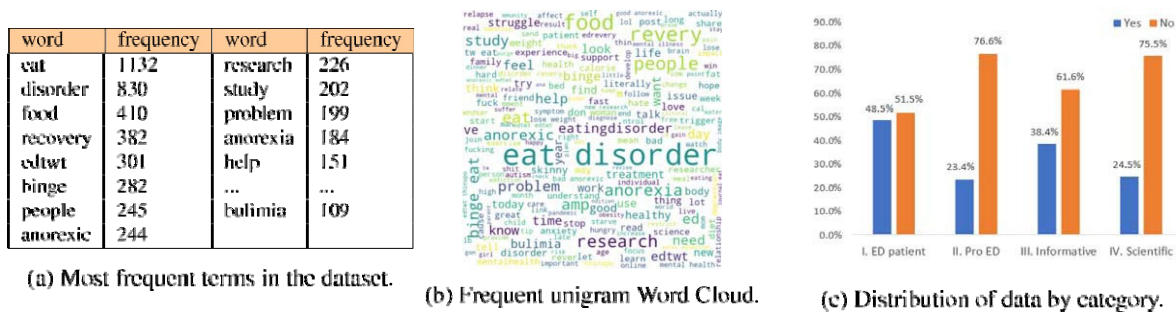

(a) Most frequent terms in the dataset.　　(b) Frequent unigram Word Cloud.　　(c) Distribution of data by category.

Fig. 5. Results of the analysis of the dataset, showing the most frequent terms and the distribution of the categories.

Table 2

Frequency of emojis in each of the four categories, and Spearman correlation between the distributions for each category

| Categorization | Frequency (label 0) | Frequency (label 1) | Spearman results |
| --- | --- | --- | --- |
| ED I | 111 | 248 | $r(163) = 0.85, p < 0.001$ |
| ED II | 231 | 128 | $r(163) = 0.96, p < 0.001$ |
| ED III | 285 | 74 | $r(163) = 0.57, p < 0.001$ |
| ED IV | 326 | 33 | $r(163) = 0.82, p < 0.001$ |

Table 3

Number of entities, unique entities, and unique entities appearing two or more times obtained from Wikidata for each of the datasets used

| Dataset | Tweets | Entities | Unique entities | Unique entities >2 |
| --- | --- | --- | --- | --- |
| Eating disorders | 2,000 | 11,680 | 1,743 | 1,358 |

* **Overlap analysis:** An overlap analysis compares the values between columns in one table or across tables. This analysis allows us to identify overlapping or redundant data in columns. In our particular case, we compared whether or not the number of emojis found in tweets that have been tagged with 0 in each of the binary categories overlapped with the emojis found that have been tagged with the value 1 in each category.
* **Spearman's rho correlation:** This analysis makes it possible to determine whether two data distributions are correlated. The analysis of emojis allows us to determine whether the distribution of emojis found in the tweets categorized in each binary class of the four categorizations performed is significantly similar or not. That is, it helps us determine whether the number of emojis found in tweets labeled as "tweets written by ED patients" is similar, statistically speaking, to the number of emojis found in tweets labeled as "tweets not written by ED patients".

It was observed that, in all cases, the similarity between the two distributions was statistically significant. The correlations are shown in Table 2. Note that in category ED III, the Spearman's rho value is lower because 8 emojis were used more than 90% of the time in tweets categorized as non-informative, compared to those categorized as informative. This difference makes sense, since, when an opinion is expressed, there are emojis that relate more to the feelings of that opinion, and informative tweets are more objective.

– Therefore, in line with what has been done in other similar research [55,66], it was decided to eliminate emojis in the pre-processing of the texts.

### 5.5. Knowledge acquisition through knowledge graph

The texts of the dataset were processed with Falcon 2.0 [58] and Entity Linker [26], resulting in a total number of entities in the Wikipedia knowledge graph, as shown in Table 3.

From the total number of entities, the total number of unique entities was calculated, and a dataset was obtained consisting of those entities that appeared at least twice in each of the tweets. By manually reviewing the entities obtained in the datasets, some disambiguation errors were detected and manually corrected (Table 4(a)).

In addition, several concepts were added to Wikidata, shown in Table 4(b). After obtaining all the entities, the knowledge graph embeddings were collected using the pyRDF2Vec tool [69] (setting *RandomWalker* with *max_depth=4* and *max_walks=50*) and the SIF algorithm is applied to combine the knowledge graph embeddings of each tweet into a single embedding. This way, a final dataset is obtained containing: the texts of the tweets, the four binary categories, and the information obtained after exploiting the knowledge graphs (knowledge graph embeddings).

### 5.6. Experiments

To validate the proposed approach, 144 tests are carried out, from which the evaluation metrics indicated above, accuracy and f1-score, are obtained. These 144 tests are divided as follows:

Table 4

Experimental tables. (a) Terms tagged with misconceptions on Wikidata by Falcon [58] and EntityLinker [26]. (b) News terms added about eating disorders added to Wikidata

| (a) Terms tagged with misconceptions on Wikidata ||| | (b) New terms added to Wikidata ||
|---|---|---|---|---|---|
| Term | Wrong link | Correct link | | Concept | Wikidata ID |
| recovery | Recovery, music album (Q274533) | Recovery approach (Q2135807) | | Food avoidance emotional disorder | Q108760799 |
| anorexia | Anorexia, music album (Q4770169) | anorexia, medical symptom (Q254327) | | fatspo | Q111780867 |
| ed | Ed, tv serie (Q930797) | eating disorder (Q373822) | | Addictive Eaters Anonymous | Q111781180 |
| binger | binger, town (Q544455) | binge eating (Q2303219) | | Meanspo | Q111781194 |
| help | The Help, a film (Q204374) | help, cooperation (Q1643184) | | Ultra-processed food | Q111781198 |

1. **Three inputs, one for each approach:** Three input datasets are applied, one for each of the approaches mentioned in this study: (i) texts, which allow obtaining the results after analyzing contextual knowledge using unstructured information; (ii) knowledge graph embeddings, which allow obtaining classifications based on contextual knowledge based on structured data; and (iii) using context-based embeddings obtained through the proposed approach.
2. **Four binary classifications:** Each machine learning model is trained four different times with each of the three inputs in order to classify the four categories mentioned in this study: ED I (tweets written by ED patient, or not), ED II (tweets promoting having an ED or not), ED III (informative tweets or not) and ED IV (scientific tweets or not).
3. **Twelve machine learning models:** The 12 models indicated in the experimental setup for the classification of the four categories indicated are run with the three different data inputs. This gives a total of 144 different outputs, 36 for each binary category to be classified, 12 with each data input in each category.

### 5.7. Dataset properties in the classification outcome

To verify that our model is not dependent on the dataset used, an analysis has been carried out on the original dataset and on the outputs obtained after applying the different models. This analysis consists of obtaining the Spearman correlation between the distributions of the most frequent terms found in tweets labeled as "0" or "1" in each category in each of the tests performed. A correlation close to 1.0 with a p-value of less than 0.05 would indicate that the distributions of the most frequent terms analyzed are statistically similar, thus being a valid and unbiased dataset for training the models. Likewise, keeping the same distribution at the output may suggest that the models do not increase or decrease the possible bias that could be present in the dataset.

After analyzing the output of the 12 models, applied on the 3 different input datasets, in the 4 tested categorizations, it has been observed that the distributions of the most frequent terms in the dataset are similar in the 144 outputs of the analyzed algorithms with a p-value < 0.001 in all cases, the maximum correlation was 1.0 and the minimum 0.996. This means that the models do not amplify the possible bias of the dataset.

### 5.8. Discussion of observed results

Twelve different classification models to classify tweets in four binary categories are compared using three different approaches: (i) contextual knowledge extracted from unstructured data, (ii) contextual knowledge extracted from structured data (KGE), and (iii) the approach proposed based in contextual knowledge extracted from structured data combined with contextual knowledge extracted from unstructured data (context-based embeddings, CBE). The results are shown in Table 5.

These results are obtained after training twelve different machine learning and deep learning models to categorize tweets in different categories using three different inputs for our dataset about eating disorders which consist of: (i) 2,000 labeled tweets, (ii) 2,000 knowledge graph embeddings and (iii) context-based embeddings obtained through our proposed approach.

Table 5

Results after applying 12 machine learning models using 3 different input variables: texts, knowledge graph embeddings, and the context-based embeddings obtained using our approach. The best results are highlighted in bold

| Model (Data) | ED I | | ED II | | ED III | | ED IV | |
|---|---|---|---|---|---|---|---|---|
| | $F_1$ | acc | $F_1$ | acc | $F_1$ | acc | $F_1$ | acc |
| RF (KGE) | 0.350 | 0.510 | 0.701 | 0.787 | 0.461 | 0.591 | 0.631 | 0.739 |
| RF (Text) | 0.774 | 0.799 | 0.910 | 0.860 | 0.808 | 0.733 | 0.921 | 0.875 |
| RF (CBE) | **0.860** | **0.860** | **0.927** | **0.878** | **0.874** | **0.843** | **0.933** | **0.902** |
| LSTM (KGE) | 0.360 | 0.499 | 0.631 | 0.787 | 0.504 | 0.602 | 0.622 | 0.739 |
| LSTM (Text) | 0.801 | 0.794 | 0.831 | 0.812 | 0.792 | 0.770 | 0.839 | 0.801 |
| LSTM (CBE) | **0.846** | **0.835** | **0.848** | **0.823** | **0.829** | **0.802** | **0.863** | **0.842** |
| Bi-LSTM (KGE) | 0.211 | 0.499 | 0.656 | 0.787 | 0.215 | 0.602 | 0.432 | 0.739 |
| Bi-LSTM (Text) | 0.791 | 0.785 | 0.841 | 0.823 | 0.777 | 0.762 | 0.864 | 0.847 |
| Bi-LSTM (CBE) | **0.835** | **0.842** | **0.853** | **0.841** | **0.835** | **0.811** | **0.893** | **0.873** |
| CNN (KGE) | 0.421 | 0.435 | 0.601 | 0.611 | 0.545 | 0.581 | 0.495 | 0.505 |
| CNN (Text) | 0.720 | 0.745 | 0.803 | 0.811 | 0.712 | 0.732 | 0.768 | 0.771 |
| CNN (CBE) | **0.821** | **0.832** | **0.821** | **0.833** | **0.805** | **0.817** | **0.834** | **0.842** |
| CNN + LSTM (KGE) | 0.490 | 0.499 | 0.490 | 0.498 | 0.314 | 0.398 | 0.421 | 0.501 |
| CNN + LSTM (Text) | 0.824 | 0.831 | 0.748 | 0.797 | 0.715 | 0.770 | 0.656 | 0.750 |
| CNN + LSTM (CBE) | **0.849** | **0.841** | **0.853** | **0.857** | **0.817** | **0.871** | **0.862** | **0.854** |
| Albert (KGE) | 0.411 | 0.552 | 0.254 | 0.782 | 0.398 | 0.580 | 0.621 | 0.739 |
| Albert (Text) | 0.852 | 0.848 | 0.930 | 0.890 | 0.872 | 0.846 | 0.944 | 0.919 |
| Albert (CBE) | **0.894** | **0.890** | **0.959** | **0.936** | **0.886** | **0.865** | **0.948** | **0.924** |
| BERT (KGE) | 0.275 | 0.526 | 0.479 | 0.796 | 0.171 | 0.586 | 0.212 | 0.648 |
| BERT (Text) | **0.876** | 0.870 | 0.933 | 0.895 | 0.887 | 0.863 | 0.951 | 0.927 |
| BERT (CBE) | 0.875 | **0.873** | **0.963** | **0.942** | **0.898** | **0.878** | **0.960** | **0.941** |
| CamemBERT (KGE) | 0.453 | 0.499 | 0.166 | 0.792 | 0.302 | 0.602 | 0.234 | 0.739 |
| CamemBERT (Text) | 0.854 | 0.841 | 0.937 | 0.900 | **0.870** | 0.846 | 0.947 | 0.924 |
| CamemBERT (CBE) | **0.885** | **0.876** | **0.961** | **0.939** | **0.870** | **0.854** | **0.950** | **0.927** |
| DistilBERT (KGE) | 0.515 | 0.577 | 0.477 | 0.786 | 0.380 | 0.551 | 0.229 | 0.670 |
| DistilBERT (Text) | 0.868 | 0.868 | 0.932 | 0.895 | 0.888 | 0.865 | 0.948 | 0.924 |
| DistilBERT (CBE) | **0.890** | **0.888** | **0.961** | **0.939** | **0.900** | **0.878** | **0.952** | **0.929** |
| FlauBERT (KGE) | 0.041 | 0.509 | 0.137 | 0.797 | 0.211 | 0.602 | 0.672 | 0.739 |
| FlauBERT (Text) | 0.852 | 0.843 | 0.928 | 0.885 | 0.848 | 0.827 | 0.949 | 0.924 |
| FlauBERT (CBE) | **0.892** | **0.888** | **0.953** | **0.927** | **0.888** | **0.868** | **0.954** | **0.932** |
| RoBERTa (KGE) | 0.499 | 0.694 | 0.350 | 0.768 | 0.211 | 0.602 | 0.234 | 0.739 |
| RoBERTa (Text) | 0.882 | 0.883 | 0.941 | 0.909 | 0.899 | 0.875 | 0.945 | 0.919 |
| RoBERTa (CBE) | **0.897** | **0.897** | **0.964** | **0.944** | **0.906** | **0.888** | **0.949** | **0.926** |
| TweetBERT (KGE) | 0.354 | 0.521 | 0.487 | 0.785 | 0.277 | 0.558 | 0.234 | 0.739 |
| TweetBERT (Text) | 0.888 | 0.887 | 0.933 | 0.898 | **0.890** | **0.868** | 0.953 | 0.931 |
| TweetBERT (CBE) | **0.909** | **0.909** | **0.964** | **0.944** | 0.888 | 0.868 | **0.957** | **0.937** |

The models trained with the context-based embeddings (CBE) dataset are the ones that obtain the best results in 97.5% of the cases. The highest percentage improvement in the $F_1$ metric occurred in ED I in the RF model, and in the acc metric occurred in the same category and model. The biggest difference in performance between the model trained with the texts and the model trained with the combination of texts and KGE in the $F_1$ metric is in the RF model in ED I, with an improvement of 11.11% in the model with texts with KGE information and in the acc metric in the RF model in ED III with an improvement of 15.01%.

There are only two cases in which the $F_1$ metric applied to the texts with KGE information data is the same or worsens with respect to that obtained in the text data, in ED III in the CamemBERT and TweetBERT models.

**Answer to RQ1.**

From the analysis of the results, it is clear that the use of contextual knowledge extracted from structured has a positive effect on the performance of the generated predictive models. Of all the experiments carried out over a dataset related to eating disorders, in 97.5% of the cases, the experiments carried out using those with knowledge graph exploitation as input data obtained better results. In the remaining 2.5% of cases, the model is equally effective or slightly less effective, not exceeding 1% worse than the model using data without KGE.

**Answer to RQ2.**

Comparing the performance results obtained by each of the seven pre-trained BERT models, the results suggest that the pre-trained model with short texts, TweetBERT, is the one that offers the best results in the metrics evaluated against its opponents in all but one occasion. These results suggest that pre-trained models with short texts may perform better than others.

**Answer to RQ3.**

It is possible to highlight that the approach proposed in this research, based on the combination of contextual knowledge obtained through structured and unstructured information, offers better results than the other two approaches studied. These results suggest that the contextual content that can be obtained from structured information, combined with the textual content of unstructured information, can help us generate more accurate predictive models. This improvement in the accuracy of models may contribute to assist some tasks of early disease detection.

## 6. Conclusions and future work

We address the problem of improving the performance of predictive models generated through machine learning techniques for short texts classification proposing a hybrid framework based on the combination of contextual knowledge from unstructured and structured data sources. Given the importance of these text classification models in different scientific and industrial contexts, any contribution that improves the performance of these predictive models is of worldwide interest and, more specifically, in the health field. Thanks to the improvement of these text classifiers, it is possible to detect and diagnose diseases earlier so that interventions can be planned to help reduce the number of patients who may have serious consequences due to these diseases.

This research presents a hybrid approach using a combination of contextual knowledge from unstructured data (e.g., BERT) and contextual knowledge from structured data (e.g., knowledge graphs), obtaining contextual knowledge embeddings. One of the objectives achieved with these embeddings is to improve classification models in short texts. It has been demonstrated on a real health-related problem over a dataset about eating disorders. The tests to validate the proposed approach show that the models trained with contextual-based embeddings have a higher success rate than those obtained with models trained only with contextual knowledge from structured or unstructured data. As a result, 97.5% of the trained models perform better with our approach than the usual ones. It is possible to highlight that using semantic information in knowledge graphs can improve the performance of predictive models in natural language processing and text mining, with the corresponding importance in the health sector. Thus, the proposed method contributes to the portfolio of tools to support understanding the disorders that social media users may suffer. Given the urgent need to help these communities, we hope these results will motivate using our proposed methods in other health-related classification problems.

Given that the approach presented in this research has been tested on a dataset within a particular health domain, eating disorders, further testing is needed to bring more robustness to the validation of the approach. In the future, we will test on a dataset with a larger amount of data to provide further validity, and we will test this approach by adding the information of emojis by combining other Knowledge Bases such as, e.g., Emojinet or Emojipedia. Finally, creating a framework that applies our approach to a given dataset is on our future agenda.

## Acknowledgements

Part of this research was funded by the European Union's Horizon 2020 research and innovation programme under Marie Sklodowska-Curie Actions (grant agreement number 860630) for the project "NoBIAS – Artificial Intelligence without Bias". This work reflects only the authors' views, and the European Research Executive Agency (REA) is not responsible for any use that may be made of the information it contains. Furthermore, Maria-Esther Vidal is partially supported by Leibniz Association in the program "Leibniz Best Minds: Programme for Women Professors", project TrustKG-Transforming Data in Trustable Insights with grant P99/2020.